\documentclass{article}

\usepackage{arxiv}

\usepackage[utf8]{inputenc} 
\usepackage[T1]{fontenc}    
\usepackage{hyperref}       
\usepackage{url}            
\usepackage{booktabs}       
\usepackage{amsfonts}       
\usepackage{nicefrac}       
\usepackage{microtype}      
\usepackage{lipsum}		

\usepackage[numbers]{natbib}
\usepackage{doi}

\usepackage{array, multirow, graphicx}
\usepackage{wrapfig}

\usepackage{caption}
\usepackage{subcaption}

\title{A Benchmark for Compositional Visual Reasoning}

\author{%
Aimen Zerroug\textsuperscript{\textnormal{1,2,3}}, 
Mohit Vaishnav\textsuperscript{\textnormal{1,2,3}}, 
Julien Colin\textsuperscript{\textnormal{2}}, 
Sebastian Musslick\textsuperscript{\textnormal{2}}, 
Thomas Serre\textsuperscript{\textnormal{1,2}}
\\ \\
   \textsuperscript{1} Artificial and Natural Intelligence Toulouse Institute, Universit\'e de Toulouse, France\\
   \textsuperscript{2} Carney Institute for Brain Science, Dpt. of Cognitive Linguistic \& Psychological Sciences \\ 
   Brown University, Providence, RI 02912\\
  \textsuperscript{3} Centre de Recherche Cerveau et Cognition, CNRS, Universit\'e de Toulouse, France \\
  \texttt{aimen\_zerroug@brown.edu}
}

\begin{document}

\date{}
\maketitle

\begin{abstract}

A fundamental component of human vision is our ability to parse complex visual scenes and judge the relations between their constituent objects. AI benchmarks for visual reasoning have driven rapid progress in recent years with state-of-the-art systems now reaching human accuracy on some of these benchmarks. Yet, a major gap remains in terms of the sample efficiency with which humans and AI systems learn new visual reasoning tasks. Humans' remarkable efficiency at learning has been at least partially attributed to their ability to harness compositionality -- such that they can efficiently take advantage of previously gained knowledge when learning new tasks. Here, we introduce a novel visual reasoning benchmark, Compositional Visual Relations (CVR), to drive progress towards the development of more data-efficient learning algorithms. We take inspiration from fluidic intelligence and non-verbal reasoning tests and describe a novel method for creating compositions of abstract rules and associated image datasets at scale. Our proposed benchmark includes measures of sample efficiency, generalization and transfer across task rules, as well as the ability to leverage compositionality. We systematically evaluate modern neural architectures and find that, surprisingly, convolutional architectures surpass transformer-based architectures across all performance measures in most data regimes. However, all computational models are a lot less data efficient compared to humans even after learning informative visual representations using self-supervision. Overall, we hope that our challenge will spur interest in the development of neural architectures that can learn to harness compositionality toward more efficient learning. 

\end{abstract}

\section{Introduction}

Visual reasoning is a rather complex ability considering the high dimensionality of the sensory input and the level of abstraction it requires. It highlights human's capacity to manipulate concepts and relations as symbols extracted from the visual input. The efficiency with which humans learn new visual concepts and relations as exemplified by fluidic intelligence and non-verbal reasoning tests is equally fascinating. In the pursuit of human-level artificial intelligence, a growing body of research is attempting to emulate this skill in machines, and deep neural networks are at the forefront of the field. 

Deep learning approaches are prime candidates as models of human intelligence due to their success in learning from data while using simple design principles. However, these architectures are still lacking due to the enormous amounts of data required for training, inability to generalize to unfamiliar situations~\cite{geirhos2020shortcut} and the lack of robustness~\cite{goodfellow2014explaining}. Their ability to perform well in large-data regimes has skewed research in the field towards scaling up datasets and architectures with little consideration for the sample efficiency of these systems. 

\begin{wrapfigure}[33]{Rt}{0.5\textwidth}
\includegraphics[width=0.5\textwidth]{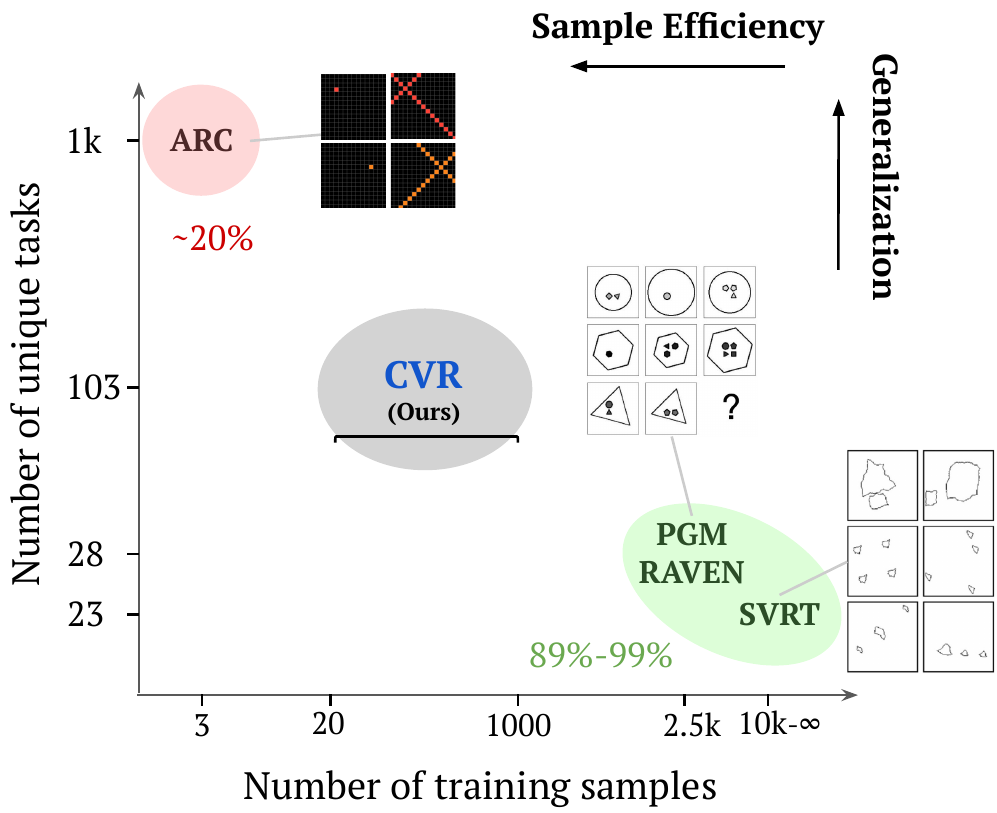} 
\caption{\textbf{Visual reasoning benchmarks}: State-of-the-art models achieve super-human accuracy~\cite{wu2020scattering, vaishnav2022understanding} on several visual-reasoning benchmarks such as RAVEN~\cite{Zhang2019-ru} PGM~\cite{barrett2018measuring} and SVRT~\cite{Fleuret2011-jm}. However, some benchmarks continue to pose a challenge for current models, such as ARC~\cite{Chollet2019-fd}. The fundamental difference between these different benchmarks is the number of unique task rules that they compose out of their priors and the number of samples that are available for training architectures on individual rules. This difference sheds light on two poorly researched aspects of human intelligence: learning in low-sample regimes and harnessing compositionality. The proposed CVR challenge aims to fill the gap between current benchmarks to encourage the development of more sample efficient and more versatile neural architectures for visual reasoning.}
\label{fig:main_fig}
\end{wrapfigure} 

Only a few benchmarks address these aspects of human intelligence. Among them, ARC~\cite{Chollet2019-fd} provides diverse visual reasoning problems. However, the extreme scarcity of training samples, only 3 samples per task, renders the benchmark difficult for all methods, especially neural networks. Other benchmarks have guided progress in the field -- helping spur the development of several neural network-based models~\cite{barrett2018measuring, Zhang2019-ru, Fleuret2011-jm}. Some focus on evaluating the task's perceptual requirements~\cite{Fleuret2011-jm}, such as detecting features, recognizing objects, perceptual grouping and spatial reasoning, while others evaluate the logical reasoning requirements~\cite{barrett2018measuring, Zhang2019-ru}, such as symbolic reasoning, making analogies and causal reasoning. However, they lack either in the variety of abstract relations incorporated in the scene or in the semantic and structural variety of scenes over which they instantiate these abstract relations.

Creating novel visual reasoning tasks can be challenging. In this benchmark, we standardize a process for creating tasks compositionally based on an elementary set of relations and abstractions. This process allows us to exploit a wide range of visual relations as well as abstract rules, thus, making it possible to evaluate both the perceptual and logical requirements of visual reasoning. Interestingly, the compositional nature of the tasks gives an opportunity to investigate the learning strategies wielded by existing methods. Among these methods, we focus on state-of-the-art abstract visual reasoning models and standard vision models. These models have been shown to reach high performance on several visual reasoning tasks in previous works~\cite{wu2020scattering, vaishnav2022understanding}, but they always require large amounts of data. A subject of interest in this paper is quantifying the sample efficiency of these models. 
\paragraph{Contributions}
Our contributions can be summarized as follows:
\vspace{-\topsep}
\begin{itemize}
      \setlength{\parskip}{0pt}
      \setlength{\itemsep}{0pt plus 1pt}
    \item A novel visual reasoning benchmark \textbf{Compositional Visual Relations} (CVR) with 103 unique task rules over distinct scene structures.
    \item A novel method for generating visual reasoning problems with a compositionality prior.
    \item A systematic analysis of the sample efficiency of baseline visual reasoning architectures.
    \item An empirical study of models' capacity at harnessing compositionality to solve complex problems.
\end{itemize}
\vspace{-\topsep}
Our large-scale experiments capture a multitude of setups including: training on joint and individual dataset tasks, pretraining with self-supervision on dataset images to contrast learning of visual representations vs. abstract visual reasoning rules, training over a range of data regimes and tests for transfer learning between dataset tasks. We present an in-depth analysis of task difficulty, which provides insights into the strengths and weaknesses of current models. Overall, we find that the best baselines trained in the most favorable conditions fall short of human sample efficiency for learning those same tasks. While models appear to be capable of transferring knowledge across tasks, we show that they do not leverage compositionality to efficiently learn task components. By releasing our dataset, we hope to inspire research on more efficient visual reasoning models. The code for generating the full dataset and training models is available \href{https://github.com/aimzer/CVR}{here}.

\section{Compositional Visual Relations Dataset}


CVR is an synthetic visual reasoning dataset that builds on prior AI benchmarks~\cite{Fleuret2011-jm,Chollet2019-fd} and on a body of cognitive science literature~\cite{Ullman1987-cz} on visual reasoning. In the following, we will describe the generation process of the dataset problems.

\paragraph{Odd-One-Out}
A sample problem consists of 4 images generated such that one of them is an outlier according to a certain rule. The goal of the task is to select the outlier. The learner is expected to test several hypotheses in order to detect the outlier. This process requires them to infer the hidden scene structure and relationships between the objects. 

\begin{wrapfigure}[23]{r}{0.48\textwidth}
\includegraphics[width=0.48\textwidth]{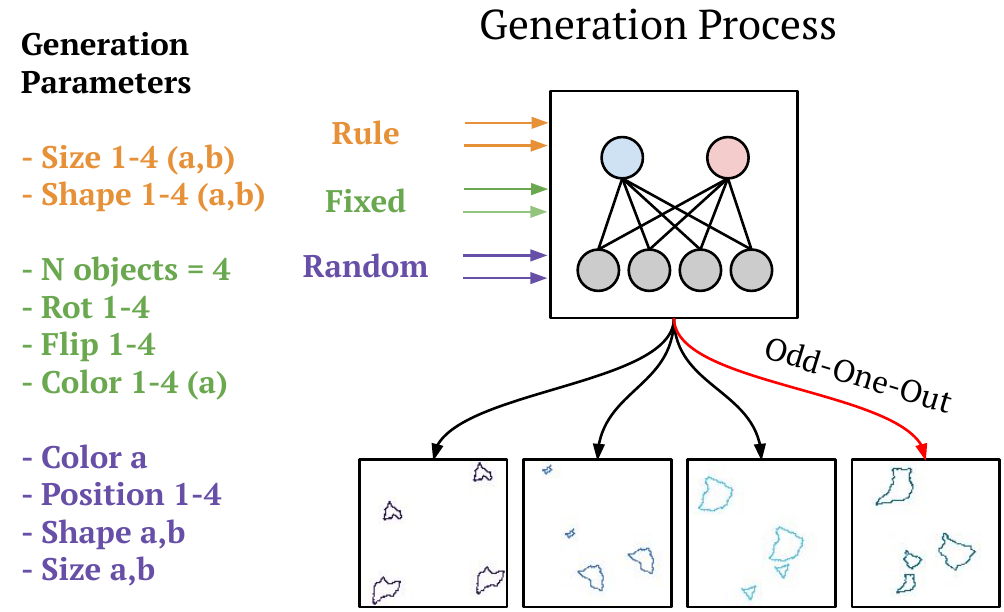}
\caption{\textbf{Odd-One-Out Generation}: Our generation process is designed to render scenes that exemplify a rule. It incorporates a reference rule and an odd-one-out rule that is designed to contradict the reference rule over target relations. Scenes are generated by varying the  input parameters of the generation process. Certain parameters are fixed across the 4 images while others are randomized. Rule-relevant parameters are fixed for 3 images (corresponding to the reference rule) but changed for the 4th image, thus making it the odd-one-out.}
\label{fig:generative_process}
\end{wrapfigure}

\paragraph{Scene generation}
Each image contains one \textbf{scene} composed of multiple \textbf{objects}. An object is defined as a closed contours with a set of \textbf{attributes}: \textit{shape}, \textit{position}, \textit{size}, \textit{color}, \textit{rotation} and \textit{flip}. Other attributes describe the scene or low-level relations between objects. \textit{Count} corresponds to the number of objects, groups of objects or relations. \textit{Insideness} indicates that an object contains another object within its contour. \textit{Contact} indicates that two object contours are touching. These 9 attributes are the basis for the 9 \textbf{elementary relations}. For example, a "size" relation is a constraint on the sizes of certain objects in the scene. Relations are expressed with natural language or logical, relational and arithmetic operators over scene attributes. Relations and objects are represented as nodes in the \textbf{scene graph}. Relations define groups of objects and can have attributes of their own. Thus, it is possible to create abstract relations over these relations' attributes. A scene can be generated from a template that we call a \textbf{structure}. The \textbf{generation process} is a program designed with a given structure to generate a scene graph and render it given a set of parameter values. A detailed example is described in Fig.~\ref{fig:generative_process}.
\begin{figure}[t!]
\centering
\includegraphics[width=1.0\textwidth]{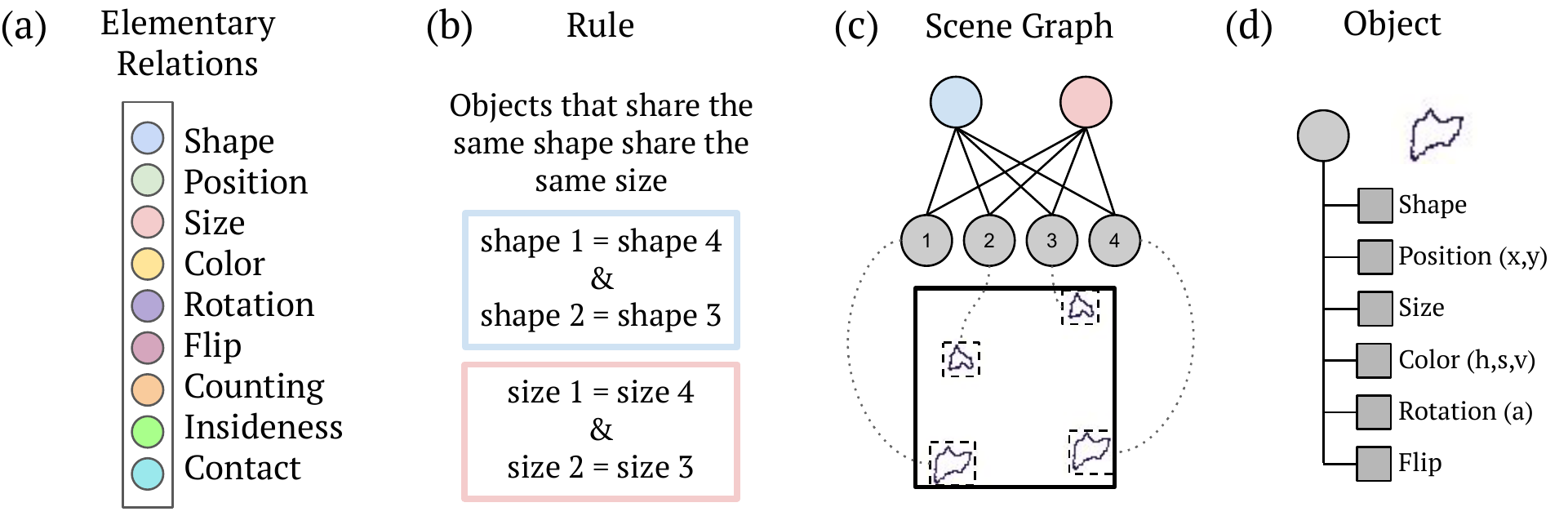}
\caption{\textbf{Scene Generation}: A scene in our image dataset is composed of objects. (d) An object is a closed contour with several attributes. (a) A relation is a constraint for the generation process over scene attributes. (b) The elementary relations control unique scene attributes and they are used for building task rules in a compositional manner. (c) Objects and relations are represented as nodes in the scene graph.}
\label{fig:dataset_description}
\end{figure}

\paragraph{Rules and problem creation}
The generation process described above can be used to instantiate different tasks; binary classification, few-shot binary classification, or a raven's progressive matrix. In this paper, we choose to apply this process to create odd-one-out problems. First, target relations are selected and incorporated into a new scene structure. In Fig.~\ref{fig:generative_process}, the target relations are size and shape similarity; they are added to a scene with 4 objects. Then, a reference rule and an odd rule are chosen such that they combine target relations in different ways. The reference and odd rules in the example vary only in the size or shape attributes. A valid odd rule cannot involve attributes irrelevant to the target relations, such as color or position in the example above. Crucially, this method ensures that any strategy employed by the learner will have to involve reasoning over the target relations in order to find the outlier. 
Given a scene structure, reference and an odd rule, the generation process has a set of free parameters that control new samples' generation. The problem's difficulty level can be varied by randomizing or fixing these parameters. A higher number of random parameters results in a higher difficulty. We create generalization test sets by changing the sets of fixed or random parameters. 

\begin{figure}[t]
\centering
\includegraphics[width=1.0\textwidth]{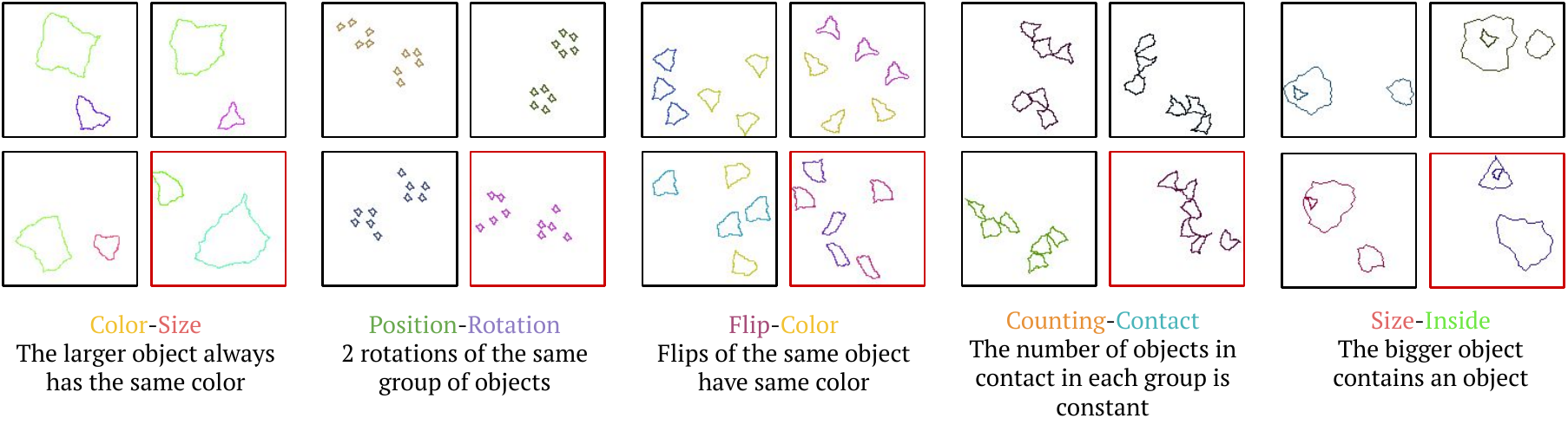}
\caption{\textbf{Examples of task rules that are composed from a pair of relations.} More examples are provided in the supplementary material.}
\label{fig:rule_examples}
\end{figure}
\paragraph{Dataset details}
CVR incorporates 103 unique reference rules, including 9 rules built on the 9 elementary visual relations and 94 additional rules built on compositions of the rules. These compositions span all pairs of elementary rules and include up to 4 relations. While certain rules are composed of the same elementary relations, they remain unique in their scene structure or associations with other relations. The procedural generation of problem samples helps us create an arbitrary number of samples. We provide 10,000 training problem samples, 500 validation samples and 1,000 test samples for each task. We also provide a generalization test set of 1000 samples.

CVR constitutes a significant extension of the Synthetic Visual Reasoning Test (SVRT)~\cite{Fleuret2011-jm} in that it provides a systematic reorganization based on an explicit compositionality prior. Among the 23 SVRT tasks, many share relations, such as tasks \#1 and \#21 which both involve shape similarity judgments. Most of these tasks can still be found amongst CVR's rules. At the same time, CVR is more general in that substituting binary classification tasks with the odd-one-out tasks allows exploring more general versions of these tasks with a broader set of task parameters. For example, in SVRT's task \#7, images of 3 groups of 2 same shapes are discriminated from images of 2 groups of 3 same shapes. This task becomes a special case in CVR of a more general \textit{shape-count} rule with $n$ groups of $m$ objects where the values are randomly sampled across problem samples. 
Unlike procedurally generated RPM benchmarks~\cite{barrett2018measuring, Zhang2019-ru}, CVR does not rely on a small set of fixed templates for the creation of task rules. The shapes are randomly created -- not chosen among a fixed set -- which renders the visual tasks difficult for models that rely on rote memorization~\cite{kim2018not} and positions are not fixed on a grid for most rules. Other attributes are sampled uniformly from a continuous interval. 

\section{Experimental setting}
\paragraph{Baseline models}
In our experiments, we select two standard vision models commonly used in computer vision. We evaluate ResNet~\cite{he2015deep}, a convolutional architecture used as a baseline in several benchmarks~\cite{barrett2018measuring, Zhang2019-ru, vaishnav2022understanding} and also used as a backbone in standard VQA models. As a transformer-based architecture, we choose ViT~\cite{dosovitskiy2020image}. It is used in various vision tasks such as image classification, object recognition, image captioning and recently in visual reasoning on SVRT~\cite{messina2021recurrent}. To evaluate the architectures on fair grounds, we choose ResNet-50 and ViT-small which have an equal number of parameters. Additionally, we evaluate two baseline visual reasoning models designed for solving RPMs: SCL~\cite{wu2020scattering} which boasts state-of-the-art accuracy on RAVEN and PGM, and WReN~\cite{barrett2018measuring} which is based on a relational reasoning model~\cite{santoro2017simple}. Finally, we present a model that combines ResNet's perception with SCL's reasoning skills. 
\paragraph{Joint vs. Individual rule learning}
Models are trained on one task or jointly on several tasks. In the context of CVR, the random generation of scenes might cause an image to be an outlier with respect to a non-target relation. For example, considering color a non-target relation, if among 4 images of an object 3 contain the same randomly sampled color and the fourth contains a different one, the fourth image can be wrongfully judged as an outlier. Thus, models trained on several tasks could easily confound rules. To avoid this problem, they are provided with a rule embedding vector. Given the rule token, models can learn different strategies for problem samples of different rules and use the correct strategy. We also compare the two settings as they allow for testing the model's capacity and efficiency at learning several strategies and routines to solve different rules. All hyperparameter choices and training details are provided in the supplementary material.
\begin{table}
  \centering
  \setlength\tabcolsep{3pt} 
    \begin{tabular}{l|l|l|rr|rr|rr|rr|rr|rr|cc}
        \toprule
        \multicolumn{3}{l|}{N train samples} & \multicolumn{2}{c|}{20} & \multicolumn{2}{c|}{50} & \multicolumn{2}{c|}{100}  & \multicolumn{2}{c|}{200}  & \multicolumn{2}{c|}{500}  & \multicolumn{2}{c|}{1000} & SES & AUC \\
        \midrule
        \parbox[t]{2mm}{\multirow{10}{*}{\rotatebox[origin=c]{90}{rand-init}}}
        & \parbox[t]{2mm}{\multirow{5}{*}{\rotatebox[origin=c]{90}{ind}}}
        & ResNet-50\cite{he2015deep}       &  28.0 &   1 &  31.1 &   1 &  32.5 &   3 &  34.0 &   6 &  38.7 &  12 &  44.8 &  24 &  33.7 &  34.9  \\
        && ViT-small\cite{dosovitskiy2020image}       &  28.6 &   1 &  30.1 &   4 &  30.9 &   4 &  31.9 &   4 &  33.8 &   4 &  35.1 &   7 &  31.3 &  31.7 \\
        && SCL\cite{wu2020scattering}           &  26.9 &   0 &  30.0 &   1 &  30.3 &   2 &  30.0 &   2 &  31.4 &   2 &  33.4 &   5 &  29.9 &  30.3  \\
        && WReN\cite{barrett2018measuring}          &  30.0 &   0 &  32.0 &   2 &  32.9 &   2 &  34.1 &   3 &  36.3 &   6 &  39.0 &  15 &  33.4 &  34.1 \\
        && SCL-ResNet 18          &  \textbf{31.4} &   1 &  \textbf{37.3} &   9 &  \textbf{37.8} &   9 &  \textbf{39.6} &  15 &  \textbf{42.7} &  21 &  \textbf{48.3} &  26 &  \textbf{38.4} &  \textbf{39.5} \\
        \cmidrule{2-17}
        & \parbox[t]{2mm}{\multirow{5}{*}{\rotatebox[origin=c]{90}{joint}}}
        & ResNet-50     &  \textbf{27.5} &   0 &  28.2 &   0 &  29.9 &   2 &  33.9 &   6 &  \textbf{52.1} &  29 &  59.2 &  34 &  36.0 &  38.4 \\
        && ViT-small     &  27.3 &   1 &  27.8 &   2 &  28.0 &   1 &  28.1 &   1 &  29.9 &   2 &  31.4 &   3 &  28.4 &  28.7 \\
        && SCL           &  25.8 &   0 &  25.8 &   0 &  28.3 &   1 &  34.1 &   3 &  43.2 &  22 &  46.2 &  27 &  32.2 &  33.9 \\
        && WReN          &  26.8 &   0 &  27.6 &   0 &  28.5 &   0 &  30.1 &   0 &  36.4 &   9 &  42.3 &  20 &  30.9 &  32.0 \\
        && SCL-ResNet 18 &  26.4 &   0 &  \textbf{28.4} &   0 &  \textbf{31.6} &   4 &  \textbf{40.7} &  13 &  51.4 &  32 &  \textbf{64.0} &  42 &  \textbf{37.6} &  \textbf{40.4} \\
        \midrule
        \parbox[t]{2mm}{\multirow{4}{*}{\rotatebox[origin=c]{90}{SSL}}}
        & \parbox[t]{2mm}{\multirow{2}{*}{\rotatebox[origin=c]{90}{ind}}}
        & ResNet-50     &  40.5 &  13 &  47.3 &  18 &  52.9 &  29 &  56.8 &  34 &  61.9 &  42 &  \textbf{67.7} &  50 &  52.4 &  54.5 \\
        && ViT-small     &  \textbf{46.7} &  16 &  \textbf{51.6} &  24 &  \textbf{54.8} &  29 &  \textbf{57.5} &  38 &  \textbf{62.0} &  44 &  65.5 &  46 &  \textbf{54.9} &  \textbf{56.4} \\
        \cmidrule{2-17}
        & \parbox[t]{2mm}{\multirow{2}{*}{\rotatebox[origin=c]{90}{joint}}}
        & ResNet-50     &  \textbf{44.3} &  16 &  \textbf{50.3} &  24 &  \textbf{55.3} &  30 &  \textbf{59.5} &  42 &  \textbf{68.9} &  49 &  \textbf{79.2} &  59 &  \textbf{57.0} &  \textbf{59.6} \\
        && ViT-small     &  39.3 &  15 &  39.5 &  13 &  40.8 &  14 &  44.1 &  16 &  53.3 &  30 &  60.7 &  41 &  44.7 &  46.3 \\
        \bottomrule
    \end{tabular}
  \caption{\textbf{Performance comparison}: For each model, we report the accuracy and number of tasks with accuracy above 80\%. SES is the Sample Efficiency Score, it favors models with high performance in low data regimes and consistent accuracy across regimes.}
  \label{tab:main_table}
%




\end{table}

\paragraph{Self-Supervised Pretraining}
Unlike humans who spend a lifetime analysing visual information, randomly initialized neural networks have no visual experience. To provide a fair comparison between humans and neural networks, we pretrain baseline models on a subset of the training data. Self-Supervised Learning (SSL) has seen a rise in popularity due to its usefulness in pretraining models on unlabelled data. In our setting, by using SSL, we aim to dissociate feature learning from abstract visual reasoning in standard vision models. We pretrained ViT-small and ResNet-50 on 1 million images from the dataset following MoCo-v3~\cite{chen2021empirical}.

\paragraph{Human Baseline}
As found in~\cite{Fleuret2011-jm}, having 21 participants solve the 9 task rules based on elementary relations and 36 randomly sampled complex task rules is sufficient to yield a reliable human baseline. We used 20 problem samples for each rule which corresponds to the lowest number of samples used for training baseline models. Each participant performed 6 different rules. More details about the behavioral experiment are provided in the supplementary material.

\section{Results}
\paragraph{Sample Efficiency}
Baseline models are trained in six data regimes ranging from 20 to 1000 training samples. All sample efficiency results are summarized in Table \ref{tab:main_table}. The random guess accuracy level in the dataset is 25\%. We observe that most randomly initialized models are slightly above the random guess accuracy after training in low data regimes. They achieve an increase in performance only when provided with more than 500 training samples. SCL-ResNet-18 followed by Resnet-50 performs the best in high data regimes, while SCL and ViT have the lowest performance in high data regimes. This result was expected, since transformer architectures generally learn better in high data regimes (millions of data points), and consistent with prior work~\cite{vaishnav2022understanding} which finds that ViTs do not learn several SVRT tasks even when trained on 100k samples. Although SCL's performance is near chance, when augmented with a strong vision backbone, Resnet-18, it achieves the best performance. This jump in performance is indicative of the two architectures' complementary roles in visual reasoning. Results show a clear positive effect for SSL pretraining on all models. 

In order to quantify sample efficiency systematically for all models, we compute the area under the curve (AUC), which corresponds to the average performance across data regimes. We also introduce \textit{Sample Efficiency Score} (SES) as an empirical evaluation metric for our experimental setting. It consists of a weighted average of accuracy where the weights are reversely proportional to number of samples: $ SES = \frac{\sum_{n}{ a_n w_{n} } }{\sum_{n}{w_n}} $ where $w_n = \frac{1}{1+log(n)}$ and $n$ is the number of samples. 
This score favors models that learn with the fewest samples while considering consistency in the overall performance. We observe that SCL-ResNet 18 scores the highest in the individual and joint training settings. In the SSL finetuning condition, ViT and Resnet-50 have a similar SES when trained on individual tasks but ResNet-50 performs better in the joint training setting. These results hint at the efficiency of convolutional architectures in visual reasoning tasks.

The best results in this framework are achieved by training baseline models in the 10k data regime jointly after SSL pretraining where ResNet-50 achieves $93.1\%$ and ViT-small achieves $81.6\%$. This high performance asserts these models' capacity to learn the majority of rules in the dataset and strongly highlights that failure in lower data regimes is explained by their inefficiency. 

\begin{wrapfigure}[42]{r}{0.35\textwidth}
\includegraphics[width=.35\textwidth]{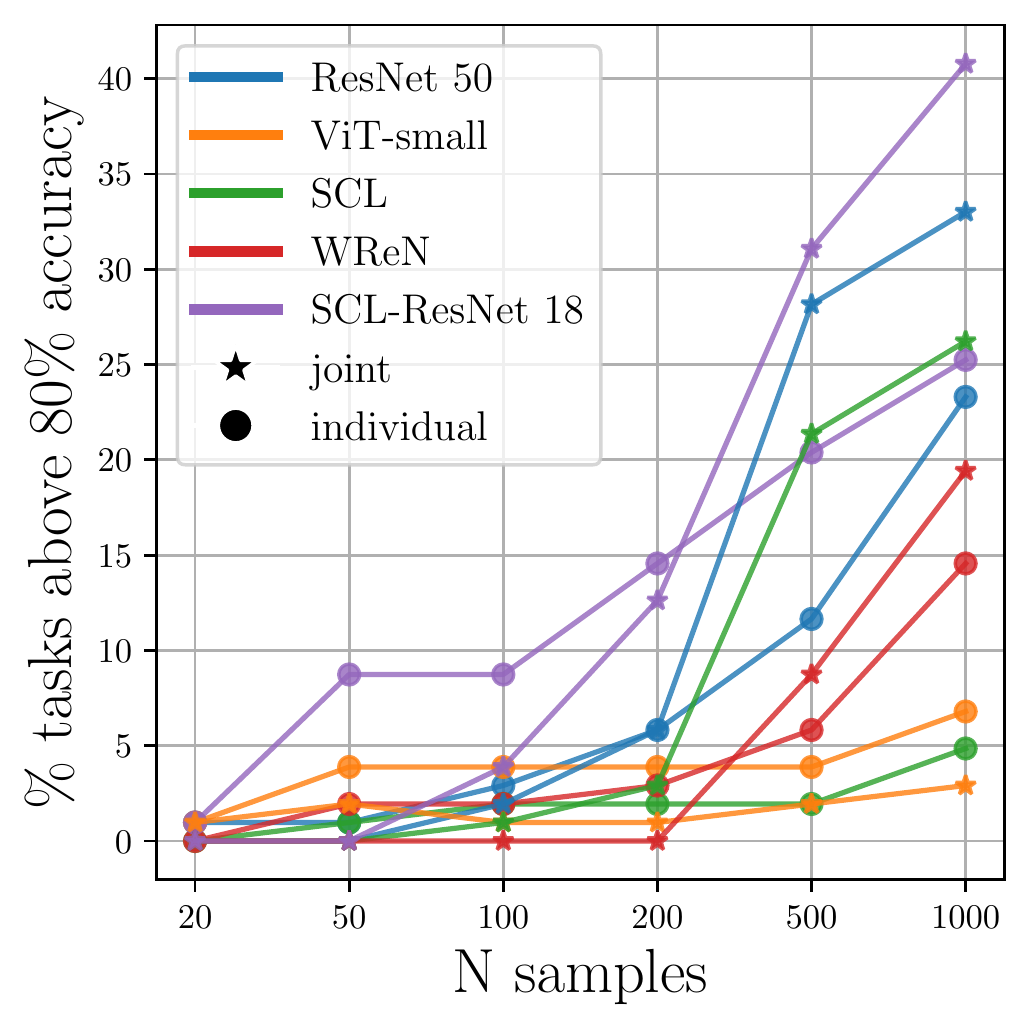}
\includegraphics[width=.35\textwidth]{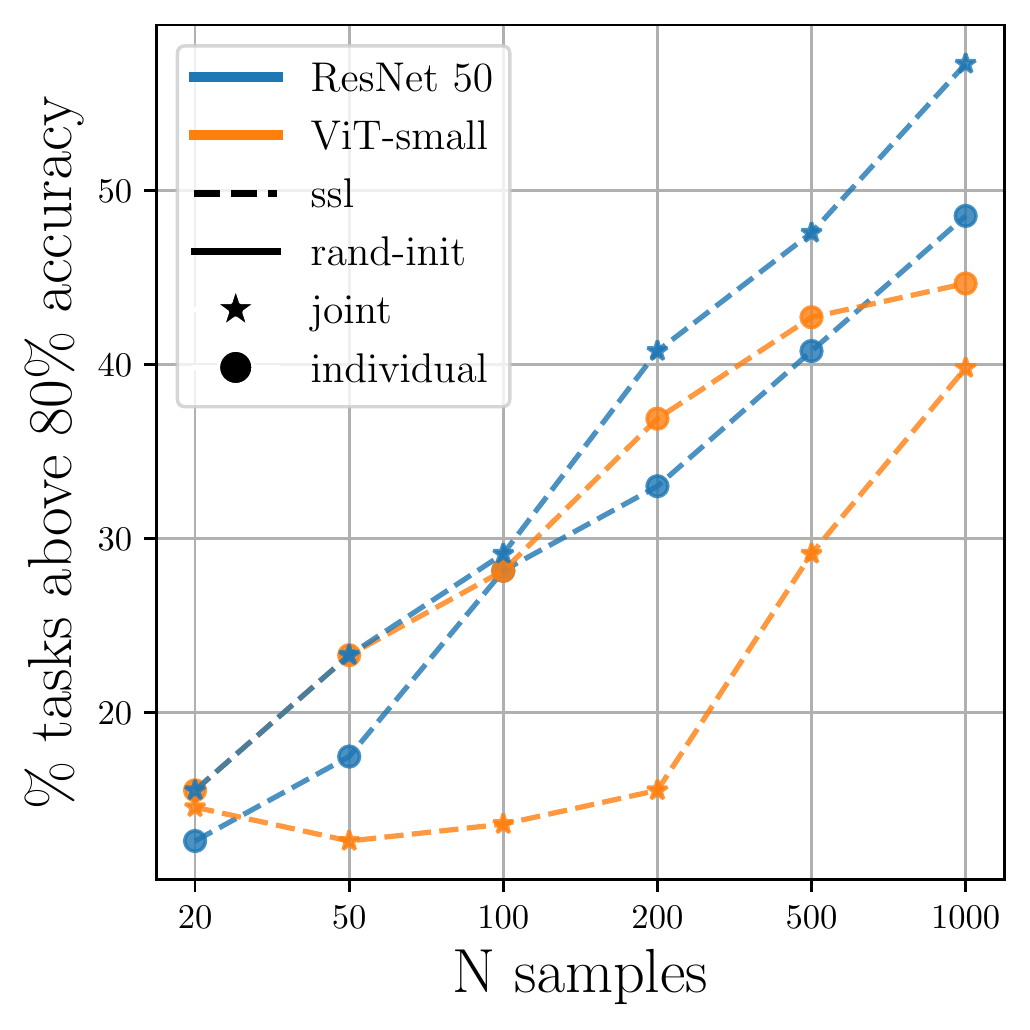}
\caption{\textbf{Sample efficiency}: The percentage of tasks for which performance is above 80\% plotted against the number of training samples per task rule, with random initialization (top) and with SSL pretraining (bottom). We observe the best performance in SSL-pretrained models. Joint training of all tasks improves performance for all models except ViT. Reasoning-oriented models consistently achieve a lower performance compared to ResNet and the SCL-ResNet combination achieves the best performance.}
\label{fig:sample_efficiency}
\end{wrapfigure}

Finally, we compare model performance to the human baseline. We observe in Table \ref{tab:human_exp} that humans far exceed the accuracy of all models with only 20 samples. This result is supported by previous work on the SVRT dataset~\cite{Fleuret2011-jm} where participants solved similar tasks with less than 20 samples. These results highlight the gap between humans and machines in sample efficiency and encourage the development of more efficient architectures. 

\paragraph{Compositionality}
Transferring knowledge and skills across tasks is a crucial feature of intelligent systems. With our experimental setup, this can be characterized in several ways. A compositional model should reuse acquired skills to learn efficiently. Thus, when it is trained on all rules jointly, it should be more sample efficient because the rules in the dataset share elementary components. In Table\ref{tab:main_table} and Figure\ref{fig:sample_efficiency}, we observe that ResNet-50 achieves higher performance on joint training compared to individual rule training while ViT has an opposite effect. The trend is consistent across data regimes and other settings. These results highlight convolutional architectures' learning efficiency compared to transformer architectures.

\begin{figure}[t]
\centering
    \begin{subfigure}[b]{0.65\textwidth}
        \includegraphics[width=\textwidth]{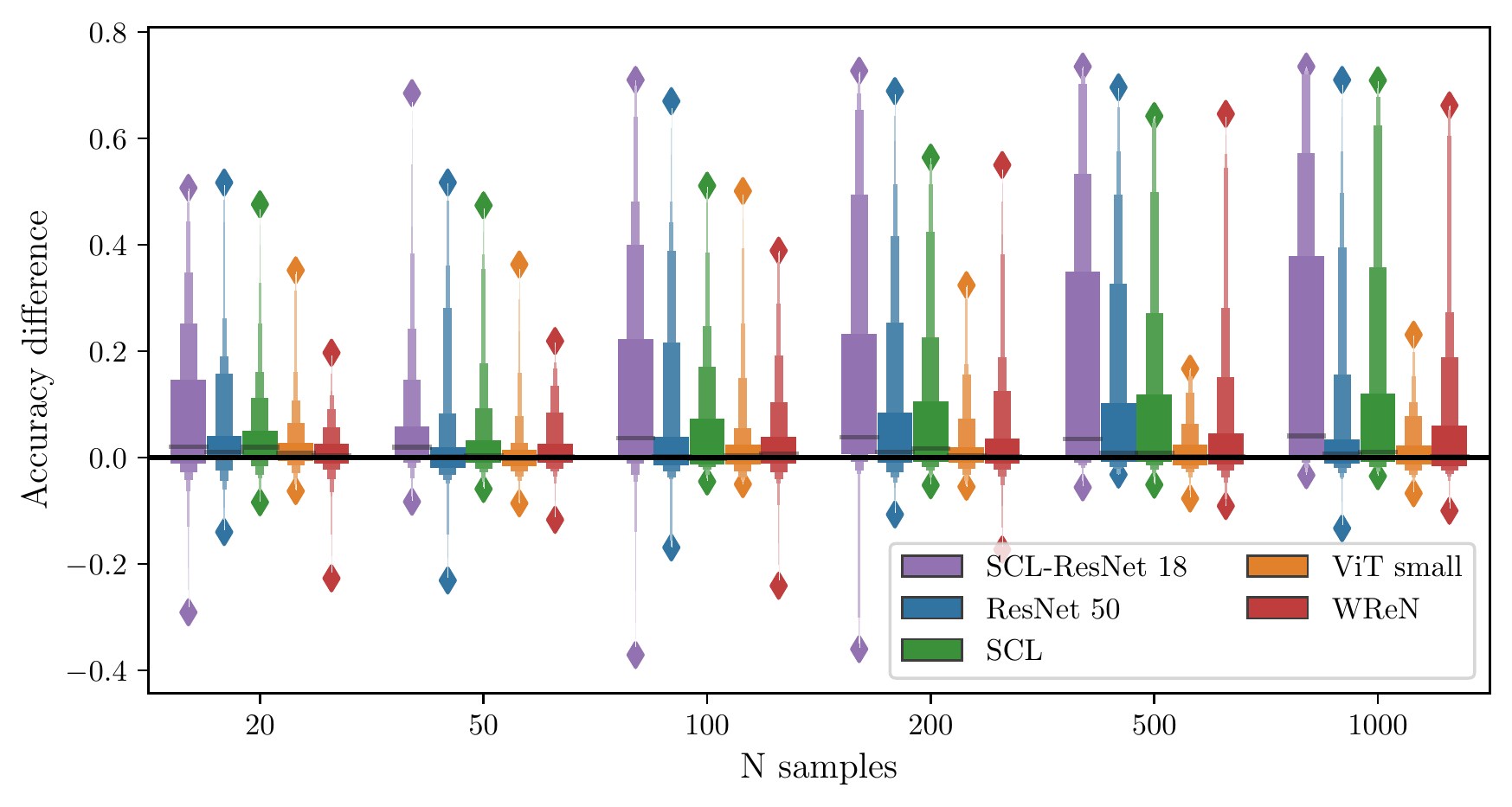}\hfill
        \caption{Curriculum}
        \label{fig:compositionality_curr_ind}
    \end{subfigure}
        \begin{subfigure}[b]{0.34\textwidth}
        \includegraphics[width=\textwidth]{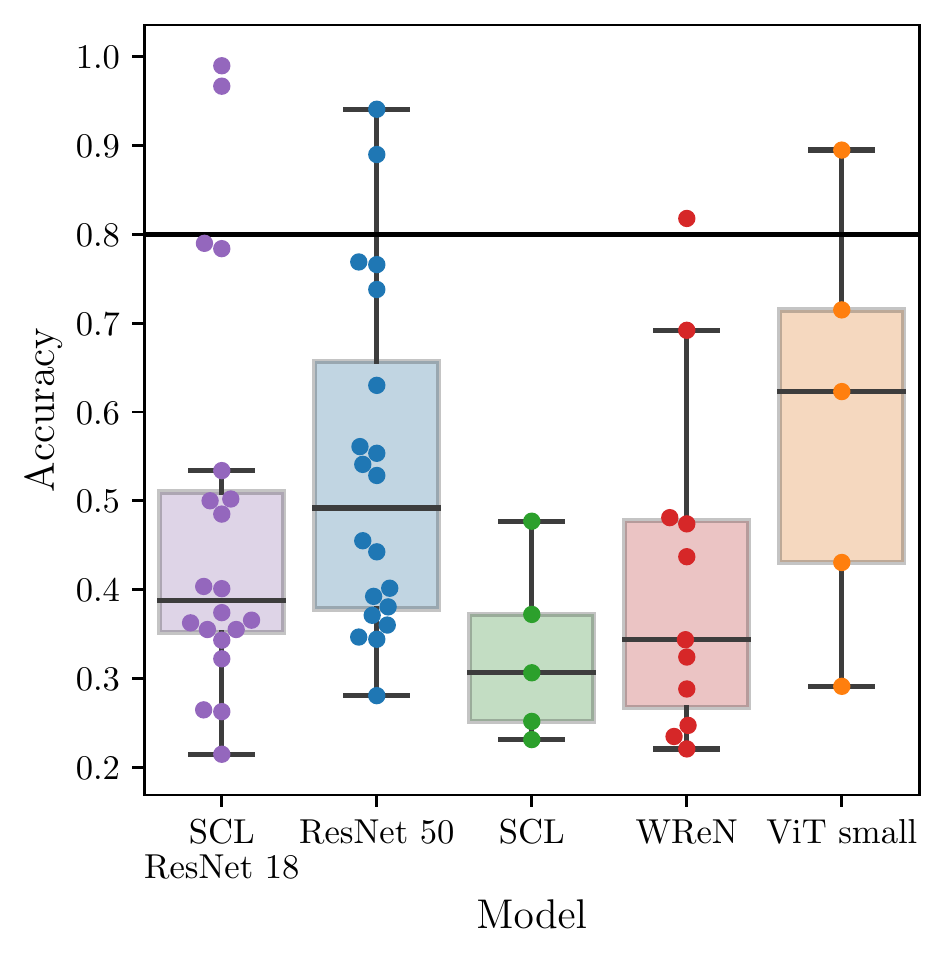}
        \caption{Reverse Curriculum}
        \label{fig:compositionality_rev_curr}
    \end{subfigure}
\caption{\textbf{Compositionality}: We evaluate models' capacity to reuse previous knowledge. 
(a) Models trained with a curriculum are compared to models trained from scratch. The distribution of difference in accuracy across tasks is plotted for each model. (b) In the 1000 samples data regime, we pick rules for which models achieved higher than 80\% accuracy and we evaluate them on the respective elementary rules.
}
\label{fig:compositionality}
\end{figure}

We investigate compositionality further by asking whether learning elementary rules provides a better initialization for learning their compositions. For example, a model that can judge object positions and sizes should not require many training samples to judge objects' relative positions identified by their sizes. We pick a set of rules with at least two different components, train models to reach the maximum accuracy possible on component relations, then finetune the models on the compositions. We call this experimental condition the curriculum condition since the condition is akin to incrementally teaching routines to a model. Model performance in the curriculum condition is compared to performance when trained from scratch. We plot the accuracy change in Fig.~\ref{fig:compositionality_curr_ind} and observe positive effects for most models but more significantly for convolution-based architectures. These results indicate that the baselines use skills acquired during pretraining to learn the composition rule to varying degrees. We refer the readers to the supplementary material for more results.

Finally, we evaluate transfer learning from composition rules to elementary rules. We name this condition the reverse curriculum condition. The working hypothesis is that models that rely on compositionality will be able to solve elementary relations without finetuning if they learn the composition. We select composition rules for which accuracy is higher than 80\% and display model accuracy on the elementary tasks in Fig.~\ref{fig:compositionality_rev_curr}. We observe that both models perform worse on the elementary relations. These results might indicate that although the baselines could transfer skills from elementary rules to their compositions, they do not necessarily use an efficient strategy that decomposes tasks into their elementary components.  

\begin{figure}[t]
\centering

    \includegraphics[width=0.20\textwidth]{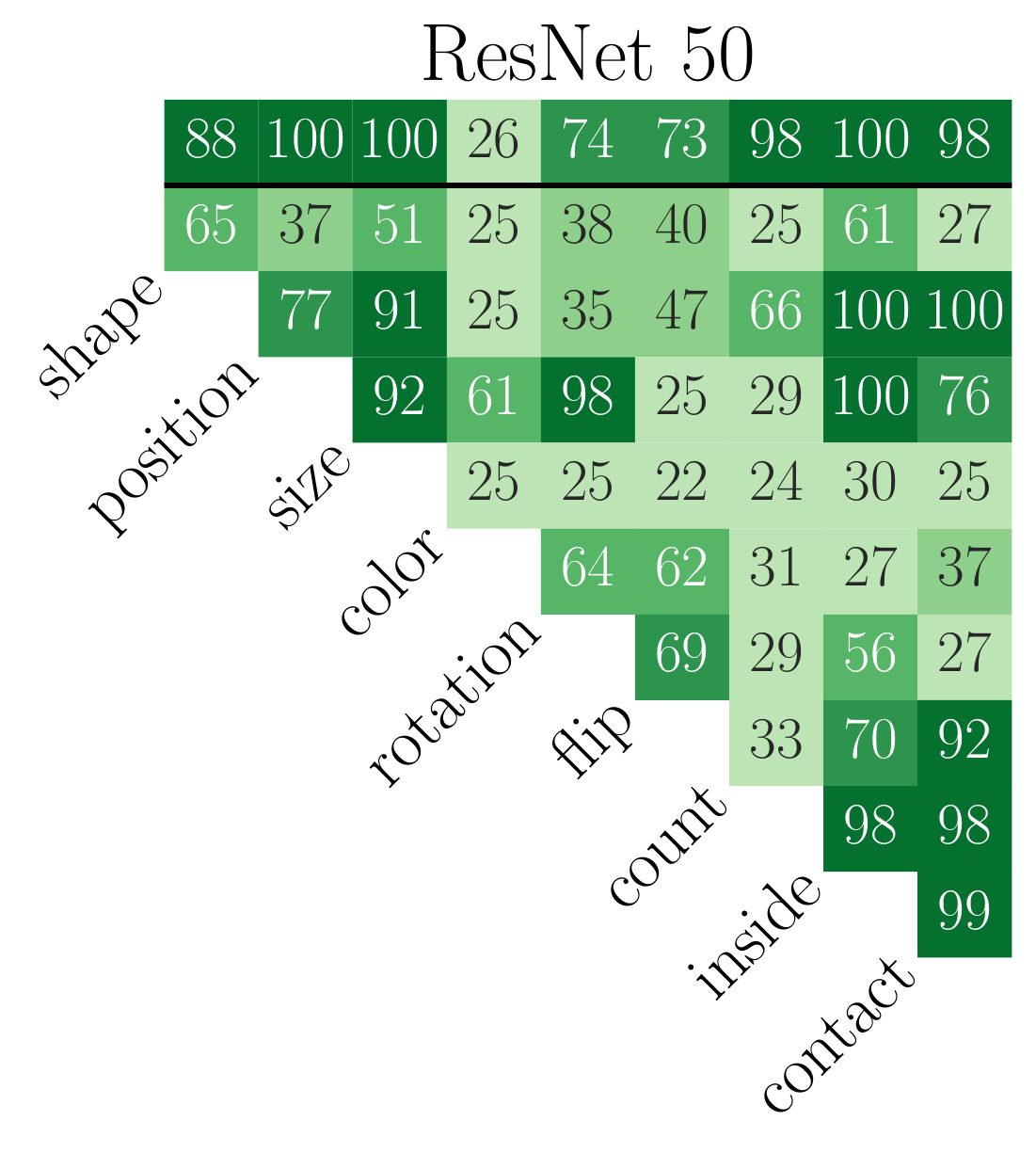}\hfill
    \includegraphics[width=0.20\textwidth]{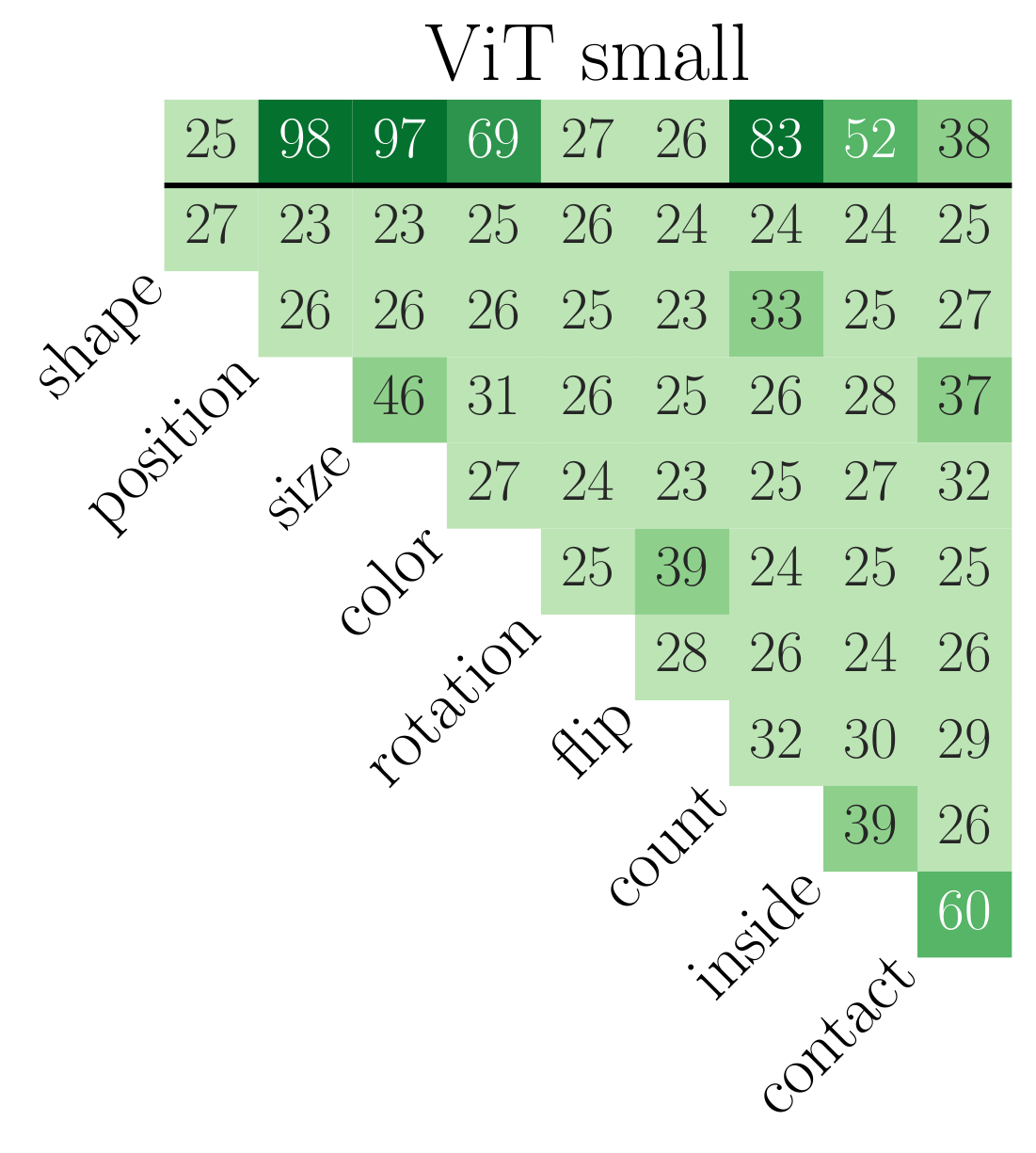}\hfill
    \includegraphics[width=0.20\textwidth]{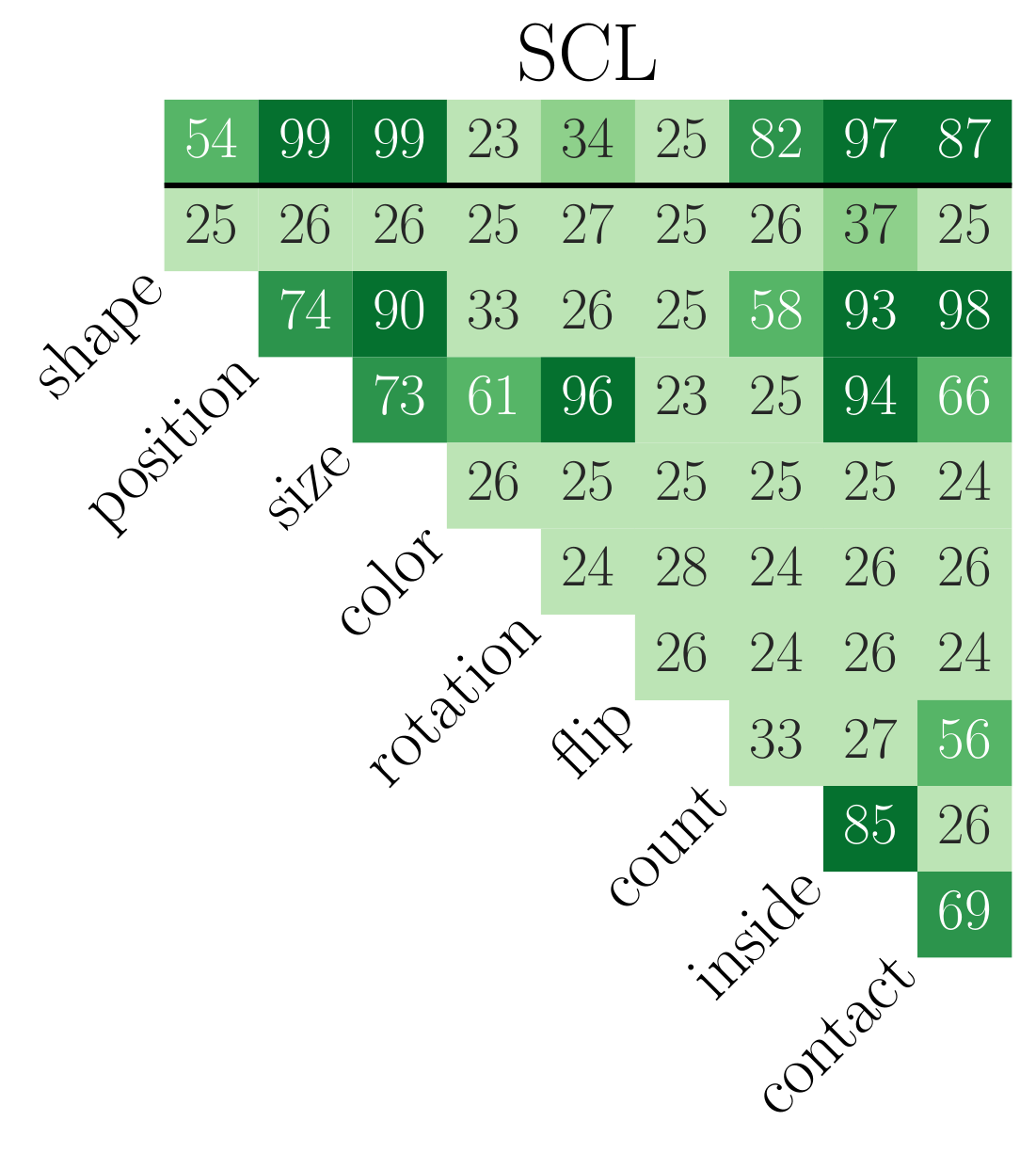}\hfill
    \includegraphics[width=0.20\textwidth]{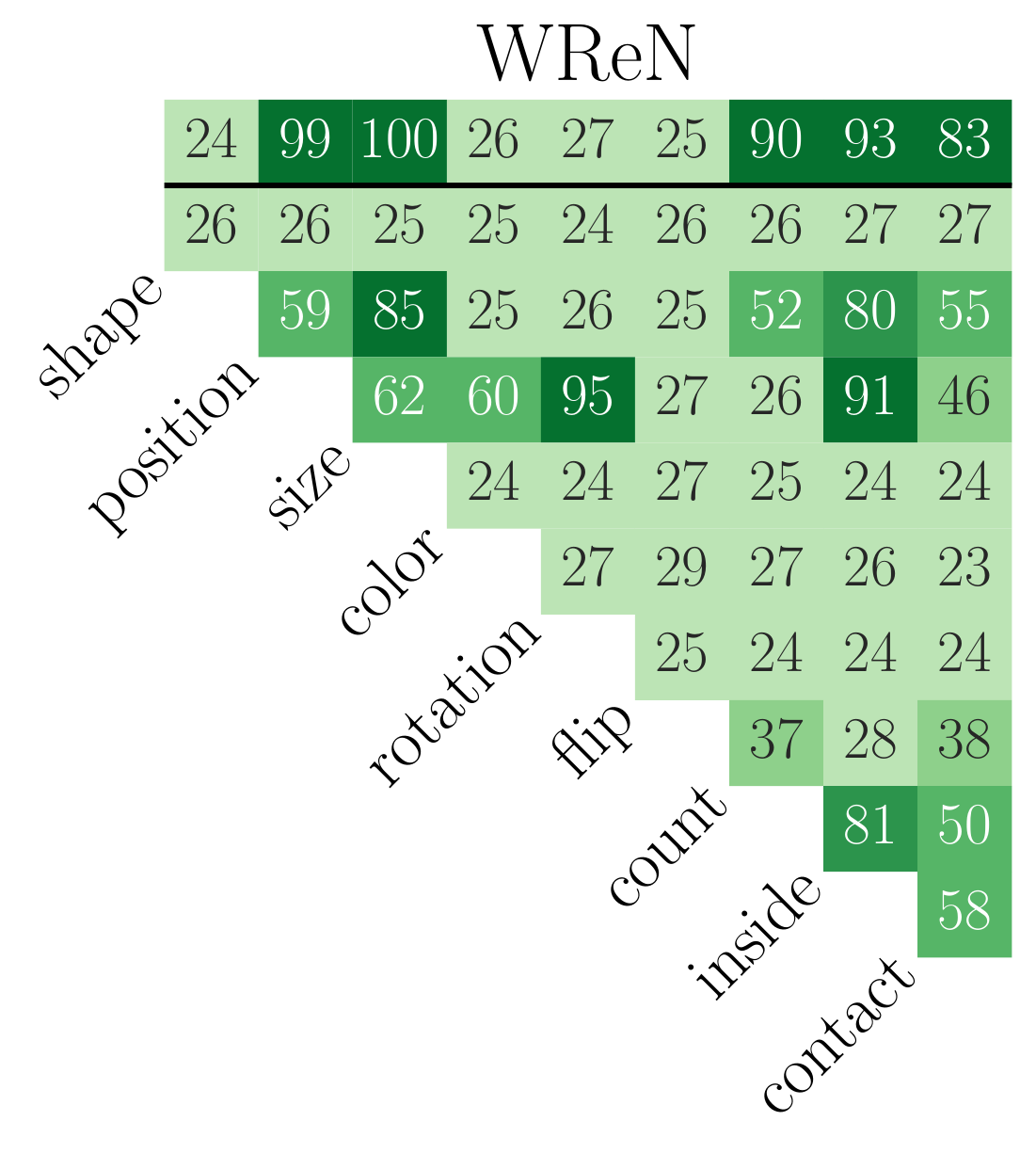}\hfill
    \includegraphics[width=0.20\textwidth]{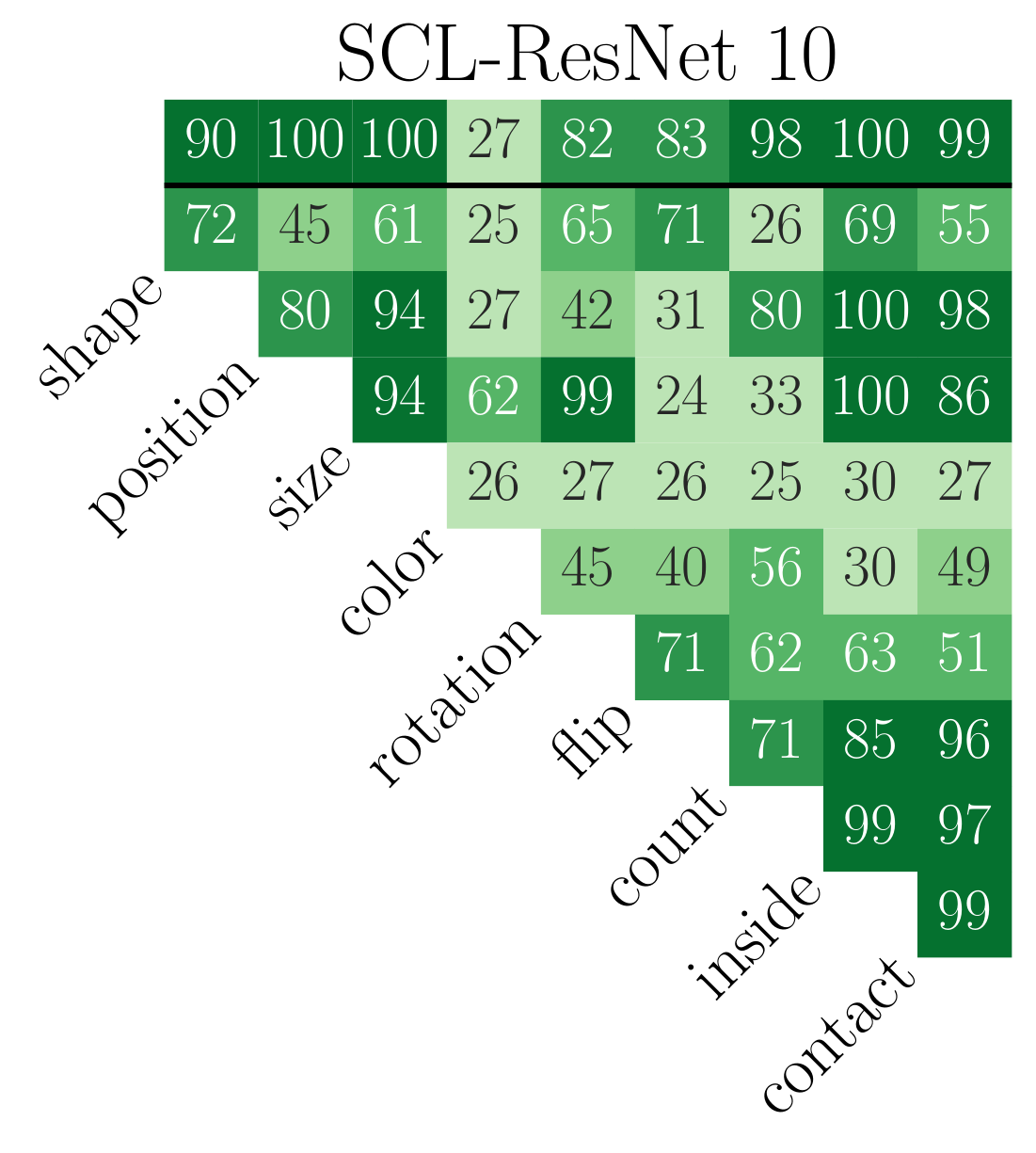}\hfill

\caption{\textbf{Task rule differences} The performance at 1000 samples is shown for each model in the joint training setting. Each square shows performance for compositions of each pair of elementary relations. Performance on elementary rules is shown on the top row. 
}
\label{fig:task_diff}
\end{figure}

\paragraph{Task difficulty}
We analyse the performance of all models in the standard setting; joint training on all rules from a random initialization. Fig.~\ref{fig:task_diff} shows the average performance of each model on each elementary rule and each group of composition rules based on the same pair. Certain rules are solvable by all models, such as the \textit{position}, \textit{size}, \textit{color}, and \textit{count} elementary rules. Additionally, certain rules pose a challenge for all models, these rules are compositions of \textit{count}, \textit{flip}, \textit{rotation} or \textit{shape}.
Models that rely on a convolutional backbone were able to solve most spatial rules; \textit{position}, \textit{size}, \textit{inside} and \textit{contact}. However, they fail on rules that incorporate shapes and their transformations; \textit{shape}, \textit{rotation}, \textit{flip}. Composition rules built with the \textit{Count} relation proved to be a challenge for most models. We believe that models are capable of solving several tasks, such as the \textit{counting} elementary rule, by relying on shortcuts; this could be a summation of all pixels in the image, for example. These shortcuts prevent models from learning abstract rules and hinder generalization. In line with the previous results, SCL-ResNet-18 seems to solve more elementary rules and compositions than the other 3 models. We believe that ViT is not capable of learning \textit{shape} based tasks because of its input processing which splits an image into patches. Since shapes are closed contours without texture in CVR, it is challenging for ViT to detect and group them.

\begin{wraptable}[22]{r}{0.44\textwidth}
    \centering
    \setlength\tabcolsep{3pt} 
    \begin{tabular}{l|rr|rr}
        \toprule
        N training samples &&       20   &&     1000 \\
        \midrule
            ResNet-50           &   28.0 &    0 &  57.9 &  14 \\
            ViT-small           &   29.3 &    1 &  32.7 &   3 \\
            SCL                 &   26.4 &    0 &  44.9 &  11 \\
            WReN                &   27.5 &    0 &  42.4 &  10 \\
            SCL-ResNet 18       &   26.8 &    0 &  \textbf{64.1} &  \textbf{18} \\
            \midrule
            ResNet-50 SSL       &   45.7 &    7 &  \textbf{78.3} &  \textbf{25} \\
            ViT-small SSL       &   38.7 &    6 &  60.3 &  17 \\
            \midrule
            Humans              &   \textbf{78.7} & \textbf{26} &-&- \\

        \bottomrule
    \end{tabular}
  \caption{\textbf{Human Baseline}: performance of models on joint training experiments is compared to the human baseline. The analysis is restricted to the 45 tasks used for evaluating humans. 
  ResNet 50 approaches human level performance only after SSL pretraining and finetuning on all task rules with 1000 samples per rule. Which is 50 times higher than the number of samples needed by humans.}
  \label{tab:human_exp}
\end{wraptable}

\section{Related Work}

\paragraph{Visual reasoning benchmarks}
Visual reasoning has been a subject of AI research for decades and several benchmarks address many relevant tasks. This includes language-guided reasoning benchmarks such as CLEVR~\cite{johnson2017clevr}, extended in its visual composition by recent work~\cite{li2022qlevr}, physics-based reasoning and reasoning over time dynamics~\cite{yi2019clevrer, bakhtin2019phyre}. More relevant to our work are abstract visual reasoning benchmarks. Raven's Progressive Matrices (RPMs) are one such example introduced in the 1938~\cite{burke1985raven} to test fluidic intelligence in humans. Procedural generation techniques for RPMs~\cite{wang2015automatic} enabled the creation of the PGM dataset and RAVEN~\cite{barrett2018measuring, Zhang2019-ru}. They also inspired Bongard-Logo~\cite{Nie2020-wm}, a concept learning and reasoning benchmark based on Bongard's 100 visual reasoning problems~\cite{bongard1968recognition}. Another reasoning dataset, SVRT~\cite{Fleuret2011-jm}, focuses on evaluating similarity-based judgment and spatial reasoning. Besides these synthetic datasets, real-world datasets were developed with similar task structures to Bongard-Logo and RPM~\cite{teney2020v, jiang2022bongard}. In this work, we take inspiration from SVRT and develop a more extensive set of rules with careful considerations for the choice of rules and using a novel rule generation method. Finally, Abstract Reasoning Corpus~\cite{Chollet2019-fd} is a general intelligence test introduced with a new methodology for evaluating intelligence and generalization. The numerous problems presented in this benchmark are constructed with a variety of human priors. The unique nature of the task, requiring solvers to generate the answer, and the limited amount of training data render the benchmark difficult for neural network based methods. We follow a similar approach in our dataset by creating several unique problem templates. However, we restrict the number of samples to a reasonable range to evaluate the sample efficiency of candidate models.

\paragraph{Neuroscience/Psychology}
Several theories attempt to propose an understanding of the mechanisms behind visual reasoning, such as gestalt psychology which outlines principles hypothesized to be used by the visual system as an initial set of abstractions. Another theory describes visual reasoning as a sequence of elemental operations called visual routines~\cite{Ullman1987-cz} orchestrated by higher-level cognitive processes. These elemental operations are hypothesized to form the basis for spatial reasoning, same-different judgment, perceptual grouping, contour tracing and many other visual skills~\cite{cavanagh2011visual}. Evaluating these skills in standard vision models is a recurring subject in machine learning and neuroscience research~\cite{kim2019disentangling}~\cite{linsley2020recurrent}~\cite{Puebla2021-tl}. To provide a comprehensive evaluation of visual reasoning, it is important to include task sets that require various visual skills within humans' capabilities. 

\section{Discussion and Future Work}
In this work, we have proposed a novel benchmark that focuses on two important aspects of human intelligence -- compositionality and sample efficiency. Inspired by visual cognition theories~\cite{Ullman1987-cz}, the proposed challenge addresses the limitations of existing benchmarks in the following ways: (1) it extends previous benchmarks by providing a variety of visual reasoning tasks that vary in relations and scene structures, (2) all tasks in the benchmark were designed with a compositionality prior, and (3) it provides a quantitative measure of sample efficiency. 

Using this benchmark, we performed an analysis of the sample efficiency of existing machine learning models and their ability to harness  compositionality. Our results suggest that even the best pretrained neural architectures require {\bf 50 times} more training samples than humans to reach the same level of accuracy, which is consistent with prior work on sample efficiency~\cite{lake2015human}. Our evaluation further revealed that current neural architectures fail to learn several tasks even when provided an abundance of samples and extensive prior visual experience. These results highlight the importance of developing more data-efficient and vision-oriented neural architectures toward achieving human-level artificial intelligence. In addition, we evaluated models' generalization ability across rules -- from elementary rules to compositions and vice versa. We find that convolutional architectures benefit from learning all visual reasoning tasks jointly and transfer skills learned during training on elementary rules. However, they also failed to generalize systematically from compositions to their individual rules. These results indicate that convolutional architectures are capable of transferring skills across tasks but do not learn by decomposing a visual task into its elementary components.

While our work addresses important questions on sample efficiency and compositionality, we note a few possible limitations of our proposed benchmark. CVR is quite extensive in terms of the visual relations it contains but it can always be further improved in its use of elementary visual relations. For example, the shapes could be parametrically generated based on specific geometric features. Hopefully, CVR can be expanded in future work to test more routines by including additional relations borrowed from other, more narrow challenges, including occlusion~\cite{kim2019disentangling}, line tracing~\cite{linsley2018learning}, and physics-based relations. The rules in the current benchmark are limited to 2 or 3 levels of abstraction to evaluate relations systematically. Other benchmarks, such as ARC\cite{Chollet2019-fd}, use higher levels of abstraction but this benchmark has proven beyond the reach of current systems. Similarly, our evaluation methods for sample efficiency and compositionality could be further improved and adapted to different settings. For example, the sample efficiency score is an empirical metric used only for evaluating our benchmark and it requires training all models on all data regimes for the score to be consistent. We hope that the release of our benchmark will encourage researchers in the field to test their own model's sample efficiency and compositionality. 

\newpage

\section{Acknowledgments}
This work is funded by  NSF (IIS-1912280) and ONR (N00014-19-1-2029) to TS. Additional support was provided by the ANR-3IA Artificial and Natural Intelligence Toulouse Institute (ANR-19-PI3A-0004). Computing resources used supported by the Center for Computation and Visualization (NIH S10OD025181) at Brown and CALMIP supercomputing center (Grant 2016-p20019, 2016-p22041) at Federal University of Toulouse Midi-Pyrénées. 

{\small
\bibliographystyle{plain}
\bibliography{references}
}

\end{document}